\documentclass[conference]{IEEEtran}
\IEEEoverridecommandlockouts
\usepackage{cite}
\usepackage{amsmath,amssymb,amsfonts}
\usepackage{algorithmic}
\usepackage{graphicx}
\usepackage{booktabs}
\usepackage{textcomp}
\usepackage{xcolor}
\def\BibTeX{{\rm B\kern-.05em{\sc i\kern-.025em b}\kern-.08em
    T\kern-.1667em\lower.7ex\hbox{E}\kern-.125emX}}
    
\begin{document}

\title{Cross-Modal Consistency-Guided Active Learning for Affective BCI Systems\\
\thanks{This research was supported by the Institute of Information \& Communications Technology Planning \& Evaluation (IITP) grant, funded by the Korea government (MSIT) (No. RS-2019-II190079, Artificial Intelligence Graduate School Program, Korea University).}
}

\author{
    \IEEEauthorblockN{Hyo-Jeong Jang}
    \IEEEauthorblockA{\textit{Dept. of Brain and Cognitive Engineering} \\
    \textit{Korea University}\\
    Seoul, Republic of Korea \\
    h\_j\_jang@korea.ac.kr}
    \and
    \IEEEauthorblockN{Hye-Bin Shin}
    \IEEEauthorblockA{\textit{Dept. of Brain and Cognitive Engineering} \\
    \textit{Korea University}\\
    Seoul, Republic of Korea \\
    hb\_shin@korea.ac.kr}
    \and
    \IEEEauthorblockN{Kang Yin}
    \IEEEauthorblockA{\textit{Dept. of Artificial Intelligence} \\
    \textit{Korea University}\\
    Seoul, Republic of Korea \\
    charles\_kang@korea.ac.kr}
}

\maketitle

%%%%%%%%%%%%%%%%%%%%%%%%%%%%%%%%%%%%%%%%%%%%%%%%%%%%%%%%%%%%%%%%%%%%%
\begin{abstract}
Deep learning models perform best with abundant, high-quality labels, yet such conditions are rarely achievable in EEG-based emotion recognition. Electroencephalogram (EEG) signals are easily corrupted by artifacts and individual variability, while emotional labels often stem from subjective and inconsistent reports—making robust affective decoding particularly difficult.
We propose an uncertainty-aware active learning framework that enhances robustness to label noise by jointly leveraging model uncertainty and cross-modal consistency. 
Instead of relying solely on EEG-based uncertainty estimates, the method evaluates cross-modal alignment to determine whether uncertainty originates from cognitive ambiguity or sensor noise. A representation alignment module embeds EEG and face features into a shared latent space, enforcing semantic coherence between modalities. Residual discrepancies are treated as noise-induced inconsistencies, and these samples are selectively queried for oracle feedback during active learning.
This feedback-driven process guides the network toward reliable, informative samples and reduces the impact of noisy labels. Experiments on the ASCERTAIN dataset examine the efficiency and robustness of ours, highlighting its potential as a data-efficient and noise-tolerant approach for EEG-based affective decoding in brain–computer interface systems.

\end{abstract}


%%%%%%%%%%%%%%%%%%%%%%%%%%%%%%%%%%%%%%%%%%%%%%%%%%%%%%%%%%%%%%%%%%%%%
\section{INTRODUCTION}
Deep learning has achieved remarkable success across various domains, largely owing to the availability of large-scale, high-quality datasets that enable neural networks to learn rich and generalizable representations~\cite{DL2021, hand1995, nn1997}. However, such an abundance of clean data is rarely achievable in real-world scenarios, particularly for electroencephalogram (EEG) signals, which are inherently noisy, and highly sensitive to recording conditions and individual variability~\cite{review_paradigms, eeg2020}. Despite these challenges, EEG remains one of the most accessible and informative modalities for studying human intention and affective states, as it captures neural activity directly related to affective and perceptual processes~\cite{neurograsp, aBCI}. 
Consequently, EEG-based emotion recognition has emerged as a key research frontier in affective neuroscience and brain–computer interface (BCI) systems, demanding models that can extract stable affective representations from inherently noisy, subject-dependent neural signals~\cite{BCI2014}.

To build affect-adaptive systems capable of understanding and responding to the inherently ambiguous and subjective nature of human affective states, recent studies have explored human-in-the-loop learning frameworks, where models collaborate with human users to continually refine their affective understanding through more multimodal feedback~\cite{iros2020, RL2018}.
Among these approaches, active learning (AL) allows the model to directly query ambiguous samples from the user, who serves as the oracle, and to focus labeling on the most informative data, thereby improving both learning efficiency and affective understanding~\cite{AL2024}.

In this context, we propose an uncertainty-aware active learning framework that quantitatively models cross-modal consistency and uncertainty-driven querying to achieve robust emotion decoding.
Our approach is motivated by the insight that uncertainty in EEG signals alone cannot reliably distinguish between sensor-level noise and genuine emotional ambiguity~\cite{jang2025uncertainty}. To overcome this limitation, we leverage cross-modal semantic consistency across heterogeneous representations as a reference for reliability estimation. Since multiple modalities including EEG, audio, and vision are recorded synchronously and reflect the same underlying semantics~\cite{emotion, human2006, pr2003}, discrepancies among them often point to measurement noise or representation-level inconsistency rather than true contextual ambiguity~\cite{cmcm2024}.
Among these modalities, visual cues such as facial expressions and full-body gestures~\cite{cross2012, gesture2006} provide particularly stable and interpretable affective signals, offering a complementary reference point that neural signals alone often cannot supply~\cite{Multimodal2, body2007, hand1999}. 
By leveraging these multimodal cues, the model can more effectively isolate noise-driven uncertainty in EEG and refine unreliable representations by prioritizing informative samples~\cite{aBCI2, select2007, unusual2015}.

To realize this framework, we introduce a cross-modal consistency module that embeds multimodal affective features into a shared latent space. The module enforces semantic alignment by minimizing the distance between consistent pairs and maximizing that of mismatched ones~\cite{align2024}. After alignment, samples exhibiting substantial inter-modal discrepancies are treated as unreliable candidates, from which an uncertainty-aware query strategy actively selects the most ambiguous instances for human feedback. During the querying phase, the oracle provides corrective supervision that enables the model to resolve affective ambiguity and recalibrate uncertain representations. Through this human-in-the-loop interaction, uncertainty is systematically reduced, improving label reliability and preserving semantic coherence across modalities~\cite{pr2024}. 
% As a result, the network achieves more robust and generalizable emotion recognition~.

%% main figure
\begin{figure}[t!]
\centerline{\includegraphics[scale=0.5]{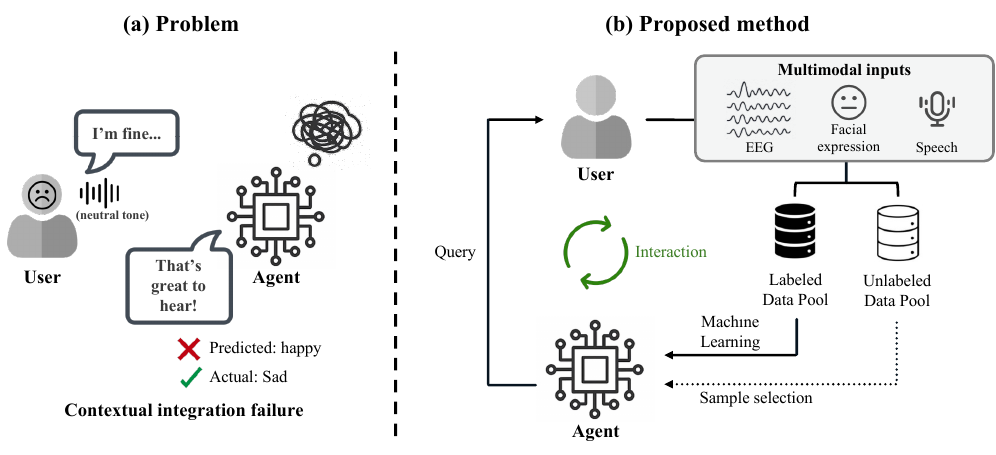}}
\caption{
Overview of the proposed framework for enhancing emotional intelligence in multimodal human–agent interaction.
(a) \textbf{Problem:} The agent with limited affective understanding misinterprets the user’s emotional states from multimodal signals, resulting in inaccurate emotion recognition and poor adaptability.
(b) \textbf{Proposed method:} The proposed framework enables the agent to improve emotional understanding via uncertainty-based active learning that selectively queries the user for feedback and updates its model.
}
\label{fig1}
\end{figure} 

In this study, we conduct experiments on the ASCERTAIN ~\cite{ascertain} benchmark to examine how ours improves labeling efficiency while maintaining robustness under noisy conditions. Overall, this study highlights the potential of uncertainty-aware active learning as a robust learning paradigm for noisy, small-scale EEG datasets in affective BCI research.

%%%%%%%%%%%%%%%%%%%%%%%%%%%%%%%%%%%%%%%%%%%%%%%%%%%%%%%%%%%%%%%%%%%%%

\section{METHODS}
\subsection{Overview}
The proposed framework integrates a multimodal consistency module and uncertainty-driven active querying into a human-in-the-loop learning paradigm.
As illustrated in Fig. 1, this framework iteratively refines affective understanding by training on labeled data, estimating uncertainty on unlabeled samples, and querying the most informative instances for annotation to progressively expand the labeled pool.
During training, it leverages multimodal signals to capture rich affective context, while at deployment, the system operates solely with EEG input, while maintaining practical deployability.

\subsection{Multimodal Consistency Module}
The system receives synchronized multimodal inputs $x = \{ x_{\text{EEG}}, x_{\text{Face}} \}$, and the goal is to infer the user's affective state $y$. The dataset is partitioned into a labeled pool $D_L = \{(x_i, y_i)\}$ and an unlabeled pool $D_U = \{x_j\}$. Training begins by optimizing the model on $D_L$. To build a coherent emotional representation across heterogeneous signals~\cite{emotion2015}, we introduce a multimodal consistency module that aligns EEG and facial embeddings within a shared latent space.
Each modality $m \in \{\text{EEG}, \text{Face}\}$ is processed by a modality- specific encoder $E_m(\cdot)$, producing a latent representation,
\begin{equation}
    z_m = E_m(x_m).
\end{equation}

To encourage consistent emotional representations across the two modalities, we initially project the latent embeddings into a shared normalized space,
\begin{equation}
    \hat{z}_m = \frac{z_m}{\| z_m \|_2},
\end{equation}
and compute a pairwise cross-modal similarity matrix,
\begin{equation}
    S(i,j) = \hat{z}_{\text{EEG}}^{(i)\top} \hat{z}_{\text{Face}}^{(j)}.
\end{equation}

The diagonal elements measure the agreement between synchronized EEG–face pairs, while the off-diagonal elements capture mismatched or redundant correlations. We enforce alignment through a symmetric contrastive loss:
\begin{equation}
    \mathcal{L}_{\text{sim}} 
    = \frac{1}{2} \Big(
        \mathrm{CE}(S, y)
        + 
        \mathrm{CE}(S^\top, y)
    \Big).
\end{equation}

While $\mathcal{L}_{\text{sim}}$ aligns paired EEG--face embeddings at the batch level, 
it treats all samples as equally reliable. 
However, affective signals frequently contain ambiguous or weakly expressive cues,
and certain samples exhibit poor cross-modal agreement even after alignment.
To account for this variability, we derive from the similarity matrix a 
sample-wise \emph{cross-modal reliability target} that quantifies how well the 
EEG embedding of each sample agrees with facial embeddings in the same batch.
For the $i$-th EEG embedding, we measure its average similarity to all facial
embeddings except its own paired sample:
\begin{equation}
    h_i = \frac{1}{N-1} \sum_{j \neq i} S(i,j).
\end{equation}

A large value of $h_i$ indicates that the EEG embedding resembles many unrelated facial embeddings. Conversely, a smaller $h_i$ implies that the sample exhibits modality-specific characteristics and is more reliable for alignment. To ensure scale invariance across batches, we normalize the scores using min--max normalization: 
\begin{equation}
    \tilde{h}_i = 
    \frac{h_i - \min(h)}{\max(h) - \min(h) + \varepsilon}.
\end{equation}

We then define the target reliability as follows, providing a stable supervisory signal:
\begin{equation}
    r_i^{\ast} = 1 - \tilde{h}_i,
\end{equation}
assigning lower reliability to samples exhibiting large cross-modal disagreement.
The model produces modality-specific reliability estimates 
$r^{\text{EEG}}_i$ and $r^{\text{Face}}_i$, which are trained to match the supervisory reliability signal:
\begin{equation}
    \mathcal{L}_{\text{rel}}
    = \frac{1}{2} \left(
        \| r^{\text{EEG}} - r^{\ast} \|_2^2
        +
        \| r^{\text{Face}} - r^{\ast} \|_2^2
    \right).
\end{equation}

While the multimodal consistency module enforces cross-modal feature alignment, the EEG head performs affective state estimation based solely on EEG embeddings, which are available at deployment. 
Given an EEG embedding $z^{EEG}$, the EEG-specific head $H(\cdot)$ predicts the affective state $\hat{y}=H(z^{EEG})$. 
For classification-based emotion recognition, the task supervision is applied through the cross-entropy loss:
\begin{equation}
    \mathcal{L}_{task} = - \sum_{c=1}^{C} y_c \log \hat{y}_c,
\end{equation}
where $C$ denotes the number of classes and $y_c$ is the target.

\subsection{Active Query Strategy}
Once the model has been optimized on the labeled pool $\mathcal{D}_L$, it is used to estimate uncertainty on the remaining unlabeled samples $\mathcal{D}_U$. The objective of the active query strategy is to identify the most informative instances whose annotation is expected to yield maximal performance improvement. Importantly, the uncertainty used for querying is derived solely from the predictive distribution of the classifier trained on multimodally aligned features.

Given an unlabeled sample $x_j \in \mathcal{D}_U$, the model produces its class probability distribution $\hat{p}_j = f(x_j)$. We quantify the uncertainty of the prediction using the Shannon entropy of the output distribution:
\begin{equation}
    \mathcal{U}(x_j)
    = - \sum_{c=1}^{C} 
        \hat{p}_j^{(c)} \log \hat{p}_j^{(c)}.
\end{equation}

Higher entropy indicates lower confidence and thus greater potential value for annotation. At each active learning iteration, all unlabeled samples are ranked by their uncertainty values. Let $\tau$ denote the acquisition ratio.  
We select the most uncertain instances by taking the top-$\tau$ fraction of the ranked set:
\begin{equation}
    \mathcal{Q}
    = \mathrm{Top}\text{-}\tau
      \big\{
        x_j \in \mathcal{D}_U \,\mid\, \mathcal{U}(x_j)
      \big\}.
\end{equation}

The queried samples $\mathcal{Q}$ are sent to the human oracle for annotation and transferred from $\mathcal{D}_U$ to the labeled pool:
\begin{align}
    \mathcal{D}_L &\leftarrow
    \mathcal{D}_L \cup \{(x_j, y_j)\}_{x_j \in \mathcal{Q}}, \\
    \mathcal{D}_U &\leftarrow
    \mathcal{D}_U \setminus \mathcal{Q}.
\end{align}

This iterative process enables the model to concentrate labeling effort on the most informative regions of the data space, resulting in a more sample-efficient gradual improvement of affective understanding.

\begin{table}[t]
\centering
\caption{Comparison of label efficiency under different training configurations.}
\begin{tabular}{lcccccc}
\toprule
\textbf{Method} & \textbf{10 \%} & \textbf{30 \%} & \textbf{50 \%} 
& \textbf{70 \%} & \textbf{100 \%} \\
\midrule
No-AL & -- & -- & -- & -- & 0.597 \\
Random & 0.462 & 0.476 & 0.526 & 0.669 & 0.812 \\
Ours (Uncertainty) & 0.462 & 0.460 & \textbf{0.584} & 0.716 & 0.853 \\
\bottomrule
\end{tabular}
\label{table:label_efficiency}
\end{table}

\subsection{Overall Objective and Training Loop}
The overall framework jointly optimizes affective representation alignment, cross-modal reliability modeling, and task-specific prediction. The total training objective integrates the three components as:
\begin{equation}
    \mathcal{L}_{\text{total}}
    =
    \lambda_{1}\, \mathcal{L}_{\text{sim}}
    +
    \lambda_{2}\, \mathcal{L}_{\text{rel}}
    +
    \lambda_{3}\, \mathcal{L}_{\text{task}},
\end{equation}
where $\lambda_{1}$, $\lambda_{2}$, and $\lambda_{3}$ act as weighting factors that regulate the relative contributions of similarity alignment, reliability regularization, and affective classification.

During each active learning iteration, the model is first optimized on the current labeled pool $\mathcal{D}_L$ using $\mathcal{L}_{\text{total}}$. The updated model then estimates uncertainty over the unlabeled pool $\mathcal{D}_U$, and the top-$p$ \% most uncertain samples are queried from the human oracle. The newly annotated instances are added to $\mathcal{D}_L$ while removed from $\mathcal{D}_U$, enabling the model to progressively refine its affective understanding.

\begin{figure}[t!]
\centerline{\includegraphics[scale=0.35]{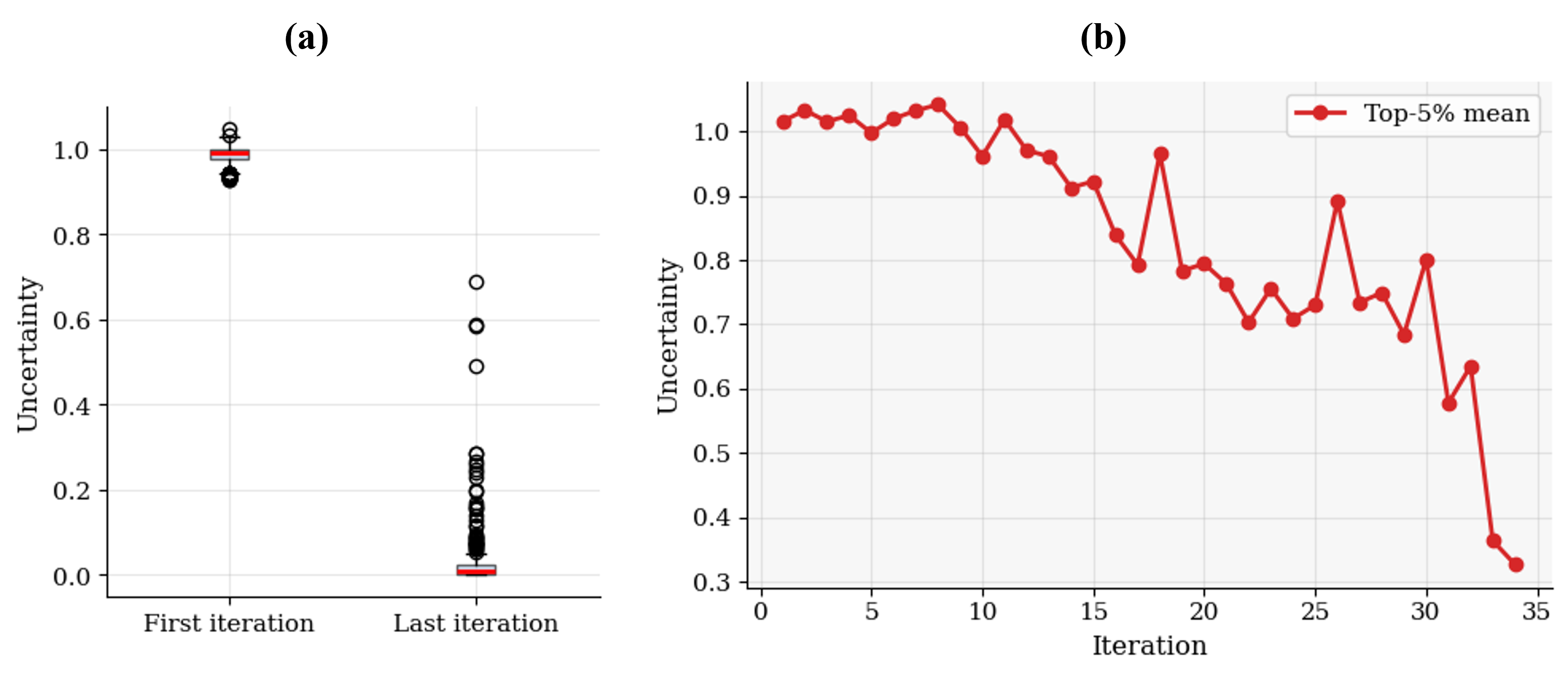}}
\caption{Evolution of uncertainty during cross-modal active learning.
(a) Comparison of uncertainty distributions between the first and last iterations.
(b) Trend of the top-5 \% uncertainty mean across active learning steps.}
\label{fig:fig_unc_score}
\end{figure}

%%%%%%%%%%%%%%%%%%%%%%%%%%%%%%%%%%%%%%%%%%%%%%%%%%%%%%%%%%%%%%%%%%%%%
\section{EXPERIMENTS}
\subsection{Experimental Setup}
Our proposed framework is evaluated using the ASCERTAIN~\cite{ascertain} dataset, a multimodal affective benchmark that jointly captures emotional states and individual personality traits under naturalistic conditions.
The dataset includes synchronized physiological and behavioral recordings such as EEG, facial expression, and audio from 58 participants viewing 36 emotional film clips.
After each clip, participants provided self-assessment scores on a 1–9 scale for \textit{valence}, \textit{arousal}, \textit{liking}, and \textit{familiarity}. 
In this work, we focus primarily on the \textit{valence} dimension, which reflects the positive–negative axis of affective experience. 
During training, 10 \% of the entire dataset is randomly selected as the initial labeled pool $\mathcal{D}_L$. Another 70 \% of the data is used as the unlabeled pool $\mathcal{D}_U$, while the remaining 20 \% is reserved as the test set.
We set $p=5$ as the querying ratio per iteration. 
All experiments are implemented in PyTorch using the Adam optimizer with a learning rate of $1\times10^{-3}$.

\subsection{Label Efficiency Across Training Configurations}
To evaluate the label-efficiency benefits of the proposed framework, we compare it against two alternative training configurations, as summarized in Table~\ref{table:label_efficiency} \textbf{No-AL} denotes a full-supervision baseline that uses 100 \% of the labeled data from the start, without multimodal consistency learning or active querying. \textbf{Random} enables the multimodal consistency module but acquires new samples uniformly at random. \textbf{Ours} represents the complete system, combining cross-modal consistency learning with uncertainty-driven sample acquisition. Across all labeling budgets, our method achieves higher accuracy while requiring substantially fewer labeled samples. Notably, at only 50 \% of the labels, our method surpasses both the Random baseline and even the fully supervised No-AL model, demonstrating the effectiveness of jointly leveraging multimodal alignment and uncertainty-aware query selection.

\subsection{Effect of Uncertainty Reduction During Active Learning}
To verify that our method effectively reduces model uncertainty over time, we examine how the predicted uncertainty scores evolve throughout the active learning loop.
Fig.~\ref{fig:fig_unc_score}(a) compares the uncertainty distribution of unlabeled samples at the beginning and the end of active learning. The first iteration exhibits high and widely spread uncertainty values, whereas the final iteration shows a sharply compressed distribution near zero, indicating that the remaining unlabeled samples have become substantially easier for the model.
Fig.~\ref{fig:fig_unc_score}(b) tracks the mean uncertainty of the
top--5 \% most uncertain samples selected at each iteration. The steady downward trend demonstrates that the queried samples become progressively less uncertain, confirming that the proposed strategy successfully reduces overall uncertainty within the pool.
In summary, this experiment validates that our uncertainty-based acquisition mechanism effectively drives down model uncertainty across iterations, ensuring that labeling efforts consistently target the most informative samples.

%%%%%%%%%%%%%%%%%%%%%%%%%%%%%%%%%%%%%%%%%%%%%%%%%%%%%%%%%%%%%%%%%%%%%
\section{CONCLUSIONS}
In this work, we introduced a cross-modal active learning framework designed to enhance affective brain–computer interface systems through multimodal consistency and uncertainty-driven querying. 
The proposed framework aligns heterogeneous affective signals such as EEG and facial expressions within a shared embedding space, mitigating modality discrepancies and improving emotional representation learning. 
By estimating model uncertainty on unlabeled samples and selectively querying the most informative instances, the system progressively expands the labeled pool with minimal annotation cost. 
Experiments on the dataset demonstrate that our method reliably identifies informative samples, achieves faster uncertainty reduction, and outperforms random querying strategies in both efficiency and convergence. 
Overall, this framework provides a principled step toward developing emotionally intelligent BCI agents capable of learning affective understanding through efficient human-in-the-loop interaction.

%%%%%%%%%%%%%%%%%%%%%%%%%%%%%%%%%%%%%%%%%%%%%%%%%%%%%%%%%%%%%%%%%%%%%
% \section{REFERENCES}

%%%%%%%%%%%%%%%%%%%%%%%%%%%%%%%%%%%%%%%%%%%%%%%%%%%%%%%%%%%%%%%%%%%%%

\bibliographystyle{IEEEtran}
\bibliography{reference}

\begin{thebibliography}{10}
\providecommand{\url}[1]{#1}
\csname url@samestyle\endcsname
\providecommand{\newblock}{\relax}
\providecommand{\bibinfo}[2]{#2}
\providecommand{\BIBentrySTDinterwordspacing}{\spaceskip=0pt\relax}
\providecommand{\BIBentryALTinterwordstretchfactor}{4}
\providecommand{\BIBentryALTinterwordspacing}{\spaceskip=\fontdimen2\font plus
\BIBentryALTinterwordstretchfactor\fontdimen3\font minus \fontdimen4\font\relax}
\providecommand{\BIBforeignlanguage}[2]{{%
\expandafter\ifx\csname l@#1\endcsname\relax
\typeout{** WARNING: IEEEtran.bst: No hyphenation pattern has been}%
\typeout{** loaded for the language `#1'. Using the pattern for}%
\typeout{** the default language instead.}%
\else
\language=\csname l@#1\endcsname
\fi
#2}}
\providecommand{\BIBdecl}{\relax}
\BIBdecl

\bibitem{DL2021}
P.~Ren \emph{et~al.}, ``A survey of deep active learning,'' \emph{ACM Comput. Surv.}, vol.~54, no.~9, pp. 1--40, 2021.

\bibitem{hand1995}
S.-W. Lee, ``Multilayer cluster neural network for totally unconstrained handwritten numeral recognition,'' \emph{Neural Netw.}, vol.~8, no.~5, pp. 783--792, 1995.

\bibitem{nn1997}
S.-W. Lee and H.-H. Song, ``A new recurrent neural-network architecture for visual pattern recognition,'' \emph{IEEE Trans. Neural Netw.}, vol.~8, no.~2, pp. 331--340, 1997.

\bibitem{review_paradigms}
R.~Abiri, S.~Borhani, E.~W. Sellers, Y.~Jiang, and X.~Zhao, ``{A comprehensive review of {EEG}-based brain-computer interface paradigms},'' \emph{J. Neural Eng.}, vol.~16, no.~1, p. 011001, 2019.

\bibitem{eeg2020}
S.~K. Prabhakar, H.~Rajaguru, and S.-W. Lee, ``{A framework for schizophrenia EEG signal classification with nature inspired optimization algorithms},'' \emph{IEEE Access}, vol.~8, pp. 39\,875--39\,897, 2020.

\bibitem{neurograsp}
J.-H. Cho, J.-H. Jeong, and S.-W. Lee, ``{NeuroGrasp: Real-time EEG classification of high-level motor imagery tasks using a dual-stage deep learning framework},'' \emph{IEEE Trans. Cybern.}, vol.~52, no.~12, pp. 13\,279--13\,292, 2021.

\bibitem{aBCI}
D.~Wu, B.-L. Lu, B.~Hu, and Z.~Zeng, ``{Affective brain-computer interfaces (aBCIs): A tutorial},'' \emph{Proc. IEEE}, vol. 111, no.~10, pp. 1314--1332, 2023.

\bibitem{BCI2014}
H.-I. Suk, S.~Fazli, J.~Mehnert, K.-R. Müller, and S.-W. Lee, ``{Predicting BCI subject performance using probabilistic spatio-temporal filters},'' \emph{PLoS One}, vol.~9, no.~2, pp. 1--15, 2014.

\bibitem{iros2020}
Z.~Wang, J.~Shi, I.~Akinola, and P.~Allen, ``{Maximizing BCI human feedback using active learning},'' in \emph{IEEE Int. Conf. Intell. Robots Syst. (IROS)}, 2020.

\bibitem{RL2018}
K.~Lee, S.-A. Kim, J.~Choi, and S.-W. Lee, ``{Deep reinforcement learning in continuous action spaces: A case study in the game of simulated curling},'' in \emph{Proc. Int. Conf. Mach. Learn. (ICML)}, 2018, pp. 2937--2946.

\bibitem{AL2024}
Y.~Xu, X.~Jiang, and D.~Wu, ``{Cross-task inconsistency based active learning (CTIAL) for emotion recognition},'' \emph{IEEE Trans. Affect. Comput.}, vol.~15, no.~3, pp. 1659--1668, 2024.

\bibitem{jang2025uncertainty}
H.-J. Jang, H.-B. Shin, and S.-W. Lee, ``Uncertainty-aware cross-modal knowledge distillation with prototype learning for multimodal brain-computer interfaces,'' \emph{arXiv preprint arXiv:2507.13092}, 2025.

\bibitem{emotion}
M.~A.~H. Akhand, M.~A. Maria, M.~A.~S. Kamal, and K.~Murase, ``{Improved EEG‑based emotion recognition through information enhancement in connectivity feature map},'' \emph{Sci. Rep.}, vol.~13, no.~1, p. 13804, 2023.

\bibitem{human2006}
M.~Ahmad and S.-W. Lee, ``Human action recognition using multi-view image sequences,'' in \emph{Proc. IEEE Int. Conf. Autom. Face Gesture Recognit. (FGR06)}, 2006, pp. 523--528.

\bibitem{pr2003}
S.-W. Lee and A.~Verri, \emph{{Pattern recognition with support vector machines: First international workshop, SVM 2002, Niagara Falls, Canada, August 10, 2002. Proceedings}}.\hskip 1em plus 0.5em minus 0.4em\relax Springer, 2003, vol. 2388.

\bibitem{cmcm2024}
Y.~Zhang \emph{et~al.}, ``{Cross-modal credibility modelling for EEG-based multimodal emotion recognition},'' \emph{J. Neural Eng.}, vol.~21, no.~2, p. 026040, 2024.

\bibitem{cross2012}
H.~Maeng, S.~Liao, H.~Kang, S.-W. Lee, and A.~K. Jain, ``{Nighttime face recognition at long distance: Cross-distance and cross-spectral matching},'' in \emph{Proc. Asian Conf. Comput. Vis. (ACCV)}, 2012, pp. 708--721.

\bibitem{gesture2006}
B.-W. Hwang, S.~Kim, and S.-W. Lee, ``A full-body gesture database for automatic gesture recognition,'' in \emph{Proc. IEEE Int. Conf. Autom. Face Gesture Recognit. (FGR06)}, 2006, pp. 243--248.

\bibitem{Multimodal2}
T.~Baltrušaitis, C.~Ahuja, and L.-P. Morency, ``{Multimodal machine learning: A survey and taxonomy},'' \emph{IEEE Trans. Pattern Anal. Mach. Intell.}, vol.~41, no.~2, pp. 423--443, 2018.

\bibitem{body2007}
H.-D. Yang and S.-W. Lee, ``{Reconstruction of 3D human body pose from stereo image sequences based on top-down learning},'' \emph{Pattern Recognit.}, vol.~40, no.~11, pp. 3120--3131, 2007.

\bibitem{hand1999}
S.-W. Lee and S.-Y. Kim, ``Integrated segmentation and recognition of handwritten numerals with cascade neural network,'' \emph{IEEE Trans. Syst., Man, Cybern., C. Appl. Rev.}, vol.~29, no.~2, pp. 285--290, 1999.

\bibitem{aBCI2}
M.~Soleymani, J.~Lichtenauer, T.~Pun, and M.~Pantic, ``{A multimodal database for affect recognition and implicit tagging},'' \emph{IEEE Trans. Affect. Comput.}, vol.~3, no.~1, pp. 42--55, 2011.

\bibitem{select2007}
M.-C. Roh, T.-Y. Kim, J.~Park, and S.-W. Lee, ``{Accurate object contour tracking based on boundary edge selection},'' \emph{Pattern Recognit.}, vol.~40, no.~3, pp. 931--943, 2007.

\bibitem{unusual2015}
D.-G. Lee, H.-I. Suk, S.-K. Park, and S.-W. Lee, ``Motion influence map for unusual human activity detection and localization in crowded scenes,'' \emph{IEEE Trans. Circuits Syst. Video Technol.}, vol.~25, no.~10, pp. 1612--1623, 2015.

\bibitem{align2024}
H.~Li, J.~Song, L.~Gao, X.~Zhu, and H.~Shen, ``{Prototype-based aleatoric uncertainty quantification for cross-modal retrieval},'' in \emph{Proc. Adv. Neural Inf. Process. Syst. (NeurIPS)}, 2024.

\bibitem{pr2024}
Q.~Hu, L.~Ji, Y.~Wang, S.~Zhao, and Z.~Lin, ``Uncertainty-driven active developmental learning,'' \emph{Pattern Recognit.}, vol. 151, p. 110384, 2024.

\bibitem{ascertain}
R.~Subramanian \emph{et~al.}, ``{ASCERTAIN: Emotion and personality recognition using commercial sensors},'' \emph{IEEE Trans. Affect. Comput.}, vol.~9, no.~2, pp. 147--160, 2016.

\bibitem{emotion2015}
J.~Kim \emph{et~al.}, ``Abstract representations of associated emotions in the human brain,'' \emph{J. Neurosci.}, vol.~35, no.~14, pp. 5655--5663, 2015.

\end{thebibliography}

\end{document}